# VST-Pose: A Velocity-Integrated Spatiotemporal Attention Network for Human WiFi Pose Estimation

Xinyu Zhang, Zhonghao Ye, Jingwei Zhang, Xiang Tian, Zhisheng Liang, and Shipeng Yu

**Abstract**— WiFi-based human pose estimation has emerged as a promising non-visual alternative approaches due to its penetrability and privacy advantages. This paper presents VST-Pose, a novel deep learning framework for accurate and continuous pose estimation using WiFi channel state information. The proposed method introduces ViSTA-Former, a spatiotemporal attention backbone with dual-stream architecture that adopts a dual-stream architecture to separately capture temporal dependencies and structural relationships among body joints. To enhance sensitivity to subtle human motions, a velocity modeling branch is integrated into the framework, which learns short-term keypoint displacement patterns and improves fine-grained motion representation. We construct a 2D pose dataset specifically designed for smart home care scenarios and demonstrate that our method achieves 92.2% accuracy on the PCK@50 metric, outperforming existing methods by 8.3% in PCK@50 on the self-collected dataset. Further evaluation on the public MMFi dataset confirms the model's robustness and effectiveness in 3D pose estimation tasks. The proposed system provides a reliable and privacy-aware solution for continuous human motion analysis in indoor environments. Our codes are available in https://github.com/CarmenQing/VST-Pose.

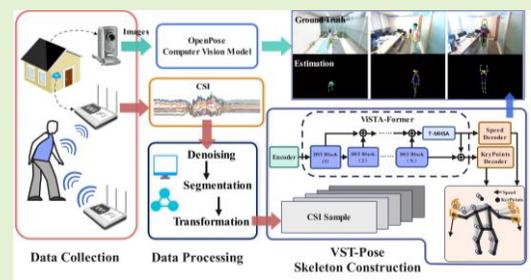

## I. Introduction

HUMAN pose estimation (HPE), which aims to reconstruct the human skeletal structure by locating multiple body joints, has been widely applied in various domains such as human-computer interaction, health monitoring, and security surveillance. Traditional HPE methods typically rely on RGB imagery [1-4] achieving high accuracy under favorable conditions. However, their performance degrades significantly in low-light environments or under occlusion due to the inherent limitations of optical sensors [5], [6] and the use of cameras raises privacy concerns in sensitive scenarios.

In recent years, WiFi-based pose estimation methods have become a popular research focus due to their good penetration, wide deployment and privacy protection advantages. Compared to earlier wireless features like Received Signal Strength Indicator or Received Signal Strength, Channel State Information (CSI) provides finer-grained and more stable measurements of the wireless channel [7]. CSI-based human sensing has demonstrated strong performance across a variety of tasks, including activity recognition [8], [9], gesture detection [10], and vital sign monitoring [11], and has driven the development of several standardized datasets, such as UT-HAR [12], NTU-HAR [13].

Compared to WiFi-based classification tasks, pose estimation demands significantly higher spatial resolution, which makes it inherently more challenging. Existing research has gradually expanded from single-person 2D [14], [15], [16] to single-person 3D [17], [18], [19] pose estimation, followed by multi-person 2D [20] and 3D [21] scenarios, as well as dense pose estimateon [22]. Additionally, recent studies have explored domain adaptation techniques to enhance cross-environment robustness [23], [24], and leveraged the Angle of Arrival to improve spatial perception [19], [25], [26].

However, most existing approaches predict a skeleton frame from a single CSI frame, failing to fully exploit the temporal continuity inherent in sequential CSI signals. This often leads to jittery and structurally inconsistent skeleton predictions in continuous pose estimation tasks. Inspired by the strong capability of Transformers [27] to model long-range dependencies, we leverage multi-head self-attention to capture temporal correlations across frames and impose spatial constraints on the structural relationships among keypoints. To further enhance model's sensitivity to subtle motion dynamics, we adopt short CSI frame sequence as input, additionally regress the velocities of keypoints, thereby improving the accuracy in HPE task.

To this end, we propose VST-Pose, a novel framework that leverages commercial off-the-shelf WiFi devices to acquire

**Contact:** Xinyu Zhang, School of Electronic and Information Engineering, South China University of Technology, China. E-mail:18010473107@163.com

CSI signals for human pose estimation. A discrete wavelet transform is applied to suppress noise in the raw CSI data. Our work adopt a supervised learning strategy, where 2D keypoint ground truth is extracted from synchronized video frames using OpenPose [4]. The overall architecture of VST-Pose comprises three key components: a convolutional neural network (CNN)-based encoder, the backbone network ViSTA-Former (Velocity-integrated Spatio-Temporal Attention Former), and a decoder.

The encoder transforms each CSI frame into a keypoint-wise embedding representation, serving as the input to the subsequent spatial attention mechanisms. ViSTA-Former, inspired by the dual-stream design of DST-Former [28], is composed of multiple cascaded Dual spatiotemporal attention blocks. Each stream within the backbone connects temporal and spatial modules in a different order to facilitate specialized modeling. Furthermore, ViSTA-Former incorporates an additional velocity modeling branch, which is fused with the keypoint branch at a late stage to enhance the estimation accuracy. The decoder maps the learned keypoint and velocity features back to the coordinate space to regress precise joint locations. Our method achieves high accuracy on both a self-constructed 2D dataset for home-care scenarios and the public 3D MMFI [29] dataset.

The main contributions of the article can be summarized as follows:

1) We develop VST-Pose, an end-to-end deep learning framework for fine-grained human pose estimation on 2.4GHz WiFi signals. The model captures the spatiotemporal dependencies between WiFi CSI and skeletal movements, enabling accurate and temporally coherent keypoint sequence estimation.

2) We introduce ViSTA-Former, a dedicated backbone network for modeling both the temporal dynamics and structural relationships of skeletons. ViSTA-Former stacks multiple Dual spatiotemporal attention blocks. and incorporates a global velocity modeling branch, equipping the network with motion-awareness and enhancing its capability to understand human movement comprehensively.

3) We conduct extensive experiments on a self-collected WiFi CSI dataset and the publicly available MMFi dataset. The results demonstrate the superior performance of VST-Pose in both 2D and 3D human pose estimation tasks.

## II. RELATED WORK

### A. Human Pose Estimation

Current mainstream HPE tasks primarily relies on visual methods that capture human keypoints from RGB images or videos. DeepPose [1] was among the earliest approaches to introduce deep learning into pose estimation, employing a top-down framework that significantly improved accuracy. Following the release of large-scale datasets, methods such as AlphaPose [2] and Cascaded Pyramid Networks [3] have also achieved state-of-the-art performance. Concurrently, bottom-up approaches exemplified by OpenPose [4] offer a favorable trade-off between accuracy and computational efficiency. Despite their high precision, visual methods tend to suffer from keypoint loss under low-light conditions and occlusions [5], [6]. Compared to vision-based methods, wearable sensor systems use inertial or angular sensors such as accelerometers and gyroscopes to track joint motion [30], offering stronger privacy protection. However, the requirement for specialized devices may limit their everyday use. LiDAR-based solutions [31] enable high-precision 3D human reconstruction via dense point clouds, though the devices are costly. Radio frequency-based methods benefiting from their excellent environmental penetration and non-intrusiveness, have emerged as important complements to vision-based techniques [32], [33]. Among these, WiFi sensing based on commercial off-the-shelf devices for acquire CSI signals shows great potential for practical deployment, owing to its low cost and widespread availability in everyday environments.

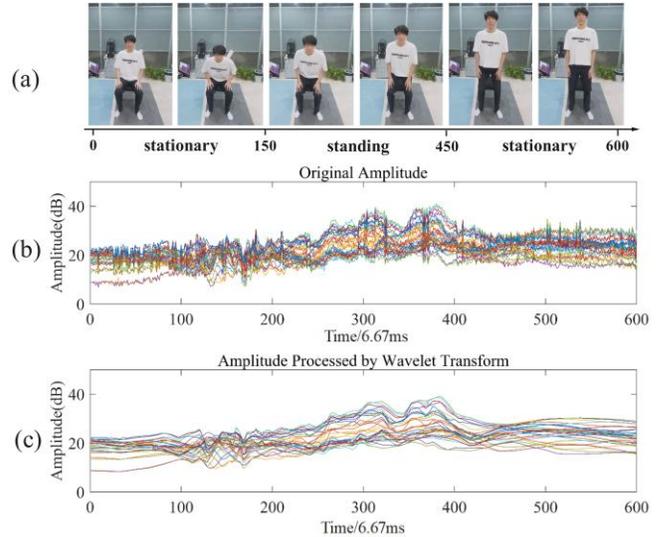

Fig. 1. 30 different colors represent the 30 subcarriers of a single antenna. (a) Shows the image of a volunteer performing the standing-up action. (b) Displays the corresponding CSI amplitude waveform at the synchronized timestamp, and (c) Shows the CSI after wavelet transform.

### B. WiFi Pose Estimation

WiFi-based human pose estimation is an emerging research area. Early works such as WiSPPN [14] employed CNNs to estimate human keypoints directly from the amplitude of CSI data frames. Guo et al. [15] further improved spatial resolution by integrating both amplitude and phase information. With the superior performance of Transformers across various domains, CSI-Former [34] stacks multiple Transformer layers as backbone to extract global feature. Metafi++ [35] leveraged attention to model spatial differences among antennas, focusing on robust target representation. Person in WiFi 3D [21] based on Transformer and the DERT framework, achieved the first multi-person 3D WiFi pose estimation. However, these approaches mainly focus on single-frame keypoint detection. In continuous pose estimation tasks, temporal modeling is critical for enhancing motion coherence and recognizing subtle movements. WiPose [17] utilized RNNs to model joint rotation angles based on a predefined skeleton topology, but was unable to distinguish human in different position. GoPose [19] combined CNNs and LSTMs but lacked explicit constraints on human body topology. To address these limitations, our work integrates multi-head attention mechanisms to model skeletal temporal correlations and structural constraints over short temporal sequences, and add velocity modeling to improve estimation accuracy for subtle motion variations.

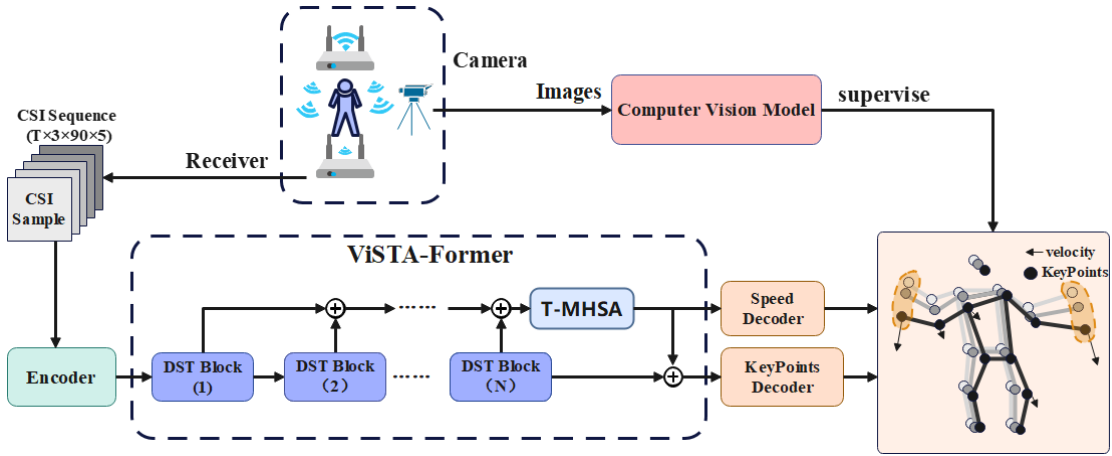

**Fig. 2.** Overview of the VST-Pose architecture. The input consists of a sequence of CSI frames from sliding windows, and the output includes both the 2D keypoint coordinates and the corresponding velocity of each keypoint.

## III. VST-Pose System Design

Our method employs two hosts equipped with Intel 5300 wireless network interface cards (NICs) operating in a master-slave configuration, working respectively as WiFi transmitter and receiver. The NICs operate on the 2.4 GHz band are each connected to three external antennas, forming a $3 \times 3$ MIMO array. In parallel, a camera is used to capture synchronized video frames for supervision.

The wireless communication is based on the IEEE 802.11n standard [36], utilizing orthogonal frequency-division multiplexing, which splits a broadband serial signal into 30 parallel narrowband subcarriers. The WiFi CSI sampling rate is 150 Hz, and the video frame rate is 30 Hz. Each subcarrier contains phase and amplitude data. Since phase data is greatly interfered by noise, we use amplitude data for pose estimation.

Fig. 1 illustrates the variation in CSI amplitude from a single antenna as a subject transitions from a seated to a standing posture. The presence of high-frequency noise superimposed on the signal may obscure the underlying patterns associated with human movement. We apply discrete wavelet transform filtering to obtain more reliable input data.

We aggregate five consecutive CSI samples into a single data frame with an initial shape of $3 \times 3 \times 30 \times 5$, representing three transmitting antennas, three receiving antennas, 30 subcarriers, and five time steps. This tensor is then reshaped into a $3 \times 90 \times 5$ CSI image, which constitutes one CSI frame for model input. Each CSI frame is temporally aligned with the corresponding RGB video frame. Ground-truth 2D keypoints $(x, y, c)$ are extracted using OpenPose, producing 25 joint coordinates. To maintain consistency with standard body representations, we retain the 17 keypoints defined in the COCO dataset, including eyes, ears, nose, neck, shoulders, elbows, wrists, hips, knees, and ankles.

## IV. Method

Our goal is to leverage the spatiotemporal dependencies in skeletons derived from CSI to enhance the network's ability to perceive human motion, enabling more accurate and continuous CSI-based human pose estimation. We propose VST-Pose, which consists of an Encoder, the ViSTA-Former subnet, and a Decoder, as illustrated in Fig. 2. The ViSTA-Former serves as the backbone, formed by cascading multiple Dual-stream Spatio-Temporal Attention (DST) blocks. Notably, we introduce an auxiliary velocity modeling branch that explicitly captures the velocity of keypoints, enhancing skeleton prediction with dynamic awareness.

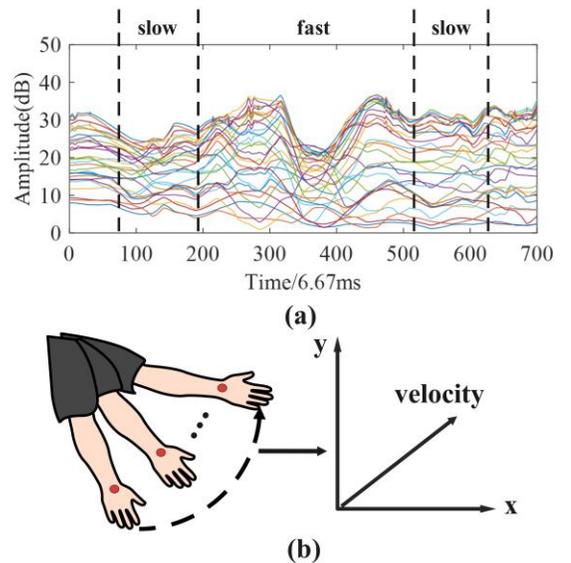

Fig. 3. **WiFi CSI velocity modeling method. (a) The impact of motion amplitude on waveform variation; (b) The calculation method for keypoint velocity.**

WiFi CSI signals enable pose estimation by sensing subcarrier variations caused by human motion. The amplitude fluctuations in the time domain are closely related to the dynamic movements of human limbs. Fig. 3 (a) illustrates the changes in CSI amplitude during a bending motion, where faster movements induce more significant waveform variations. Fig. 3 (b) depicts the velocity of the wrist keypoint over a period, calculated from the coordinate difference between the initial and final frames. This velocity includes both displacement magnitude and direction, representing the overall motion trend within that interval.

## A. CSI Encoder

The input to the network is denoted as $X \in \mathbb{R}^{T \times 3 \times 90 \times 5}$, where $T$ represents the number of consecutive CSI frames. We first apply a three-layer convolutional network to extract local spatial features from each CSI frame and obtain a coarse structural representation of human keypoints. The first convolutional layer includes a max pooling operation to downsample the spatial dimensions and extract more robust spatial patterns. In the final convolutional layer, the number of output channels is set to the number of human keypoints $J$, allowing each channel to learn the latent spatial representation of a specific keypoint through channel-wise feature extraction. The resulting feature map has a shape of $T \times J \times H \times W$. We then flatten the dimensions $H \times W$ and pass them through a fully connected layer to obtain $X_c \in \mathbb{R}^{T \times J \times D}$, where $D$ denotes the embedding dimension of each keypoint in the spatial space.

## B. ViSTA-Former: Velocity-integrated Spatio-Temporal Attention Former

Transformer-based Multi-Head Self-Attention (MHSA) mechanism introduced in Transformer [27] is effective for modeling long-range dependencies and capturing global contextual information. In our framework, we employ a Temporal Block to model the dynamic evolution of skeletons over time and a Spatial Block to capture the structural relationships among different keypoints. To incorporate positional context, we add learnable temporal positional encodings $P_T \in \mathbb{R}^{T \times 1 \times D}$ and spatial positional encodings $P_S \in \mathbb{R}^{1 \times J \times D}$ to the extracted features $X_c$. The resulting feature tensor, with shape $T \times J \times D$, serves as the input to the ViSTA-Former network.

Building upon the approach of Dual-stream Spatio-Temporal Transformer (DST-Former) [28], we adopt its dual-branch architecture, stacking Temporal and Spatial block in different orders within each branch to capture temporal relationships and structural constraints of the skeleton. On this basis, we construct the DST-Block, which further introduces a global velocity modeling branch dedicated to learning the velocity of each keypoint, thereby enhancing the model's dynamic awareness of human motion and enabling more accurate pose estimation. The network architecture of the DST-Block is illustrated in Fig. 4.

**Spatial Block.** The Spatial Block independently extracts the spatial features $F_S \in \mathbb{R}^{J \times D}$ at each CSI time step, modeling the topological relationships and relative positional information among keypoints. We adopt the Multi-Head Self-Attention (MHSA) mechanism from the Transformer, where $F_s$ is first projected into a query vector $Q$, a key vector $K$, and a value vector $V$, respectively.

$$Q^i = F_S W^{(Q,i)}, K^i = F_S W^{(K,i)}, V^i = F_S W^{(V,i)} \quad (1)$$

where $W^{(Q,i)}$, $W^{(K,i)}$, $W^{(V,i)} \in \mathbb{R}^{D \times dk}$ are the learnable linear projection matrices for the $i$-th head in MHSA, with $i \in \{1, \cdots, H\}$. $d_k$ denotes the dimensionality of each attention head.

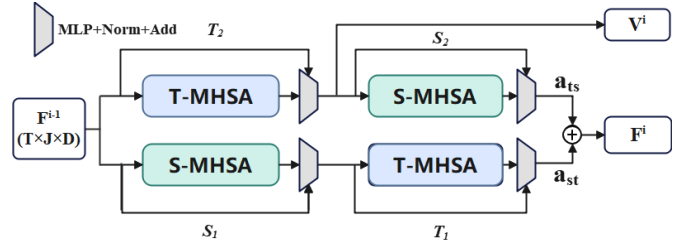

**Fig. 4.** The DST-Block structure of ViSTA-Former, consisting of two specialized attention branches, with the TS branch serving as the source of velocity modeling.

The output of the Spatial Multi-Head Self-Attention is denoted as $SMHSA \in \mathbb{R}^{J \times D}$

$$SMHSA(Q, K, V) = [head_1; \cdots; head_H]W^O,$$
$$head_i = \mathrm{softmax}\left(\frac{Q^i(K^i)'}{\sqrt{d_k}}\right)V^i \quad (2)$$

where $W^O$ is the output projection matrix, and $head_i$ represents the output of the $i$-th attention head.

The aggregated output is then passed through a Multilayer Perceptron (MLP), followed by layer normalization and a residual connection, Since the same spatial module is shared across all time steps, the final output tensor has the shape $T \times J \times D$. The entire spatial modeling process is denoted as $S(\cdot)$.

**Temporal Block.** The Temporal Block is designed to model the temporal dependencies of human skeletons in the CSI domain. We first flatten the spatial features $F_S$ at each time step into a vector of dimension $D_{flat} = J \times D$, forming the temporal sequence feature $F_T \in \mathbb{R}^{J \times D_{flat}}$. Similar to the Spatial Block, we apply Multi-Head Self-Attention to FT to obtain the temporal attention output $TMHSA \in \mathbb{R}^{J \times D_{flat}}$. This output is then passed through an MLP and layer normalization with residual connection. Finally, the result is reshaped back to the format (T,J,D). The entire temporal modeling process is denoted as $T(\cdot)$.

**Dual-Stream Spatiotemporal Fusion.** Inheriting the dual-stream modeling mechanism from DST-Former [28], each DST-Block produces two outputs: $F_i^{ST}$ from the space-first ST branch, and $F_i^{TS}$ from the time-first TS branch. The two branches do not share weights, and are indexed as 1 and 2 to represent the ST and TS branches, respectively. To effectively fuse the information from both branches, we introduce learnable weights $a_{ST}$ and $a_{TS}$. The final output of $i$-th DST-Block is denoted as $F^i \in \mathbb{R}^{T \times J \times D}$, representing the temporal skeleton feature. It is computed as:

$$F^i = a_{ST}^i \cdot T_1^i(S_1^i(F^{i-1})) + a_{TS}^i \cdot S_2^i(T_2^i(F^{i-1}))$$
$$a_{ST}^i, a_{TS}^i = \mathrm{softmax}(W(\mathrm{concat}[T_1^i(S_1^i(F^{i-1})), S_2^i(T_2^i(F^{i-1}))])) \quad (3)$$

where $i \in \{1, \cdots, N\}$ denotes the index of the $i$-th DST-Block, and $W$ is a learnable parameter vector.

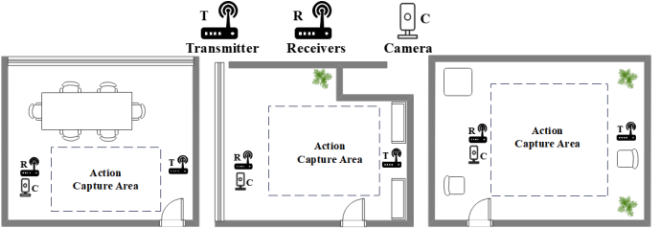

Fig. 5. Schematic of three experimental data environments. The dashed lines represent the activity range of the volunteers.

**Velocity Modeling.** We focus on modeling the velocity of keypoints within short frame sequences to capture more precise limb movements from CSI. A local velocity branch, denoted as $V^i \in \mathbb{R}^{T \times J \times D}$ is introduced in the TS branch of each DST-Block. Compared to the ST branch, endowing the TS branch with motion-awareness fully leverages its temporal modeling capabilities. Vi is computed as:

$$V^i = S_2^i(T_2^i(F^{i-1}))  \tag{4}$$

ViSTA-Former is composed of $N$ stacked DST-Blocks. We use skip connections to aggregate the local velocity features $Vi$ from all layers and feed them into a separate temporal block to obtain a global velocity feature $F_V \in \mathbb{R}^{T \times J \times D}$. A late fusion strategy is adopted to compute the final skeleton feature $F_K$, defined as $F_K = 0.5F_V + F^N$ enabling the model to capture both spatial topology and temporal dependencies while enhancing its sensitivity to dynamic changes.

### C. Decoder

The decoder maps both the keypoint backbone feature $F_K$ and the velocity $F_V$ to their corresponding shapes. and the velocity features $F_V$ to the final prediction results. Leveraging the global awareness capability of MHSA, only the final time step of $F_V$ is selected and passed through an MLP to generate the predicted global velocity vector, denoted as $V_{pred} \in \mathbb{R}^{J \times 2}$. For keypoint sequence prediction, each time step of $F_K$ is independently processed by an MLP to generate the corresponding keypoint positions, represented as $K_{pred} \in \mathbb{R}^{T \times J \times 2}$.

### D. Learning Objective

We adopt the Mean Squared Error as the loss function. Unlike previous methods that introduce smoothness terms to enforce temporal continuity of skeletons [17], our approach explicitly models the velocity of keypoints, providing more explicit motion-guided features for body movements. This enhances the model's sensitivity to subtle motion variations and avoids the motion blurring effect that may result from over-smoothing.

The total loss function $L$ is defined as a weighted combination of the keypoint position regression loss and the velocity regression loss:

$$L = \alpha \parallel V_{pred} - V_{gt} \parallel_2^2 + (1-\alpha) \parallel K_{pred} - K_{gt} \parallel_2^2,$$
$$V_{gt} = K_{gt}^T - K_{gt}^1  \tag{5}$$

where $\alpha$ is a balancing coefficient set to 0.2 in our experiments. $F_{gt}$, $V_{gt}$ denote the ground-truth of keypoint sequence and velocity, respectively. The velocity is computed as the difference between the last and first frames of the keypoint sequence.

TABLE I
THE PCK FOR EACH BODY PART ON SELF-COLLECTED DATASET.

| Keypoint | PCK@50 | PCK@40 | PCK@30 | PCK@20 | PCK@10 |
|---|---|---|---|---|---|
| Nose | 93.13 | 89.57 | 83.56 | 72.9 | 51.54 |
| L.Eye | 92.74 | 89.18 | 83.3 | 72.68 | 51.9 |
| R.Eye | 92.64 | 88.94 | 82.7 | 72.15 | 51.47 |
| L.Ear | 92.93 | 89.26 | 83.86 | 73.38 | 53.51 |
| R.Ear | 93.2 | 89.74 | 83.93 | 73.33 | 52.8 |
| L.Shoulder | 93.25 | 90.08 | 84.93 | 74.73 | 53.42 |
| R.Shoulder | 93.41 | 90.42 | 85.28 | 75.75 | 53.52 |
| L.Elbow | 91.09 | 87.88 | 82.05 | 65.7 | 36.15 |
| R.Elbow | 90.96 | 87.76 | 81.8 | 65.07 | 35.01 |
| L.Wrist | 87.03 | 81.14 | 71.48 | 59.53 | 30.8 |
| R.Wrist | 87.25 | 81.48 | 72.18 | 60.66 | 32 |
| L.Hip | 95.35 | 92.91 | 88.9 | 79.28 | 58.87 |
| R.Hip | 95.64 | 93.14 | 88.64 | 79.99 | 59.01 |
| L.Knee | 92.6 | 89.77 | 84.71 | 74.86 | 53.04 |
| R.Knee | 92.78 | 89.97 | 84.97 | 75.35 | 53.16 |
| L.Ankle | 91.93 | 88.57 | 83.54 | 73.84 | 51.86 |
| R.Ankle | 92.34 | 89.22 | 84.03 | 74.68 | 52.56 |
| Average | 92.25 | 88.79 | 82.94 | 71.9 | 48.91 |

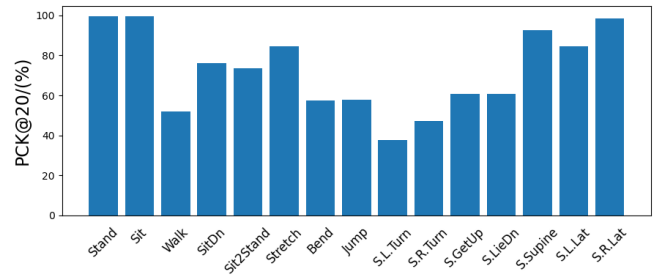

Fig. 6. The PCK@20 for 15 actions on the self-collected dataset. "S" denotes sleep-related actions, "L" denotes left, and "R" denotes right.

## V. EXPERIMENTS

### A. Set Up

We recruited five volunteers with heights ranging from 165 cm to 190 cm aged 23-32 years to participate in our data collection. Each subject was instructed to perform each designated action at least 12 times. Each action segment was trimmed to a 3-second duration corresponding to 450 CSI samples at 150 Hz, corresponding to approximately 90 frames of CSI data. To better align with the realistic needs of smart home scenarios, data collection was conducted in three distinct residential environments, as illustrated in Fig. 5. The WiFi transmitter, receiver, and camera were positioned in the same relative location and diraction. Participants were asked to move systematically across predefined areas to cover the entire activity space.

A total of 15 common in-home human actions were selected for the pose estimation task: (1) daily activities, including standing, sitting, walking, sitting down, standing up, stretching, bending, and jumping; and (2) sleep-related actions, including turning left, turning right, getting up, lying down, supine, left lateral, and right lateral positions.

The final dataset used for keypoint estimation consists of 3,300 action segments and approximately 297,000 CSI frames, each annotated with 2D keypoints following the COCO-17 standard.

During the data partitioning phase, we adopted a frame-level data augmentation strategy. Each action sequence was divided

### TABLE II
#### PERFORMANCE COMPARISON ON SELF-COLLECTED DATASET

| Method | PCK@50 | PCK@20 | MPJPE | PA-MPJPE |
|---|---|---|---|---|
| MetaFi++ [37] | 86.37 | 56.87 | 11.5248 | 6.4905 |
| Hpeli [39] | 85.60 | 51.70 | 11.9218 | 6.4454 |
| DT-Pose [24] | 88.24 | 63.15 | 10.5659 | 5.7397 |
| Ours | **92.25** | **71.90** | **8.0692** | **4.7016** |

### TABLE III
#### PERFORMANCE COMPARISON ON MMFI DATASET

| Method | PCK@50 | PCK@20 | MPJPE | PA-MPJPE |
|---|---|---|---|---|
| MetaFi++ | 85.93 | 47.71 | 187.9 | 106.5 |
| Hpeli | 86.38 | 49.52 | 183.0 | **105.3** |
| DT-Pose | 86.29 | 49.53 | 183.3 | 105.6 |
| Ours | **87.37** | **55.03** | **169.7** | 105.9 |

into 10 data clips to encourage the model to learn the spatial variations of the same action. Each clip contains 9 consecutive CSI frames and is fed into the short-term temporal modeling network using a sliding window mechanism. In total, we obtained 33,200 CSI segments, which were randomly split into training and testing sets with a ratio of 4:1. This segmentation strategy ensures compatibility with both single-frame and multi-frame models and provides a fair comparison baseline.

To further evaluate the 3D human pose estimation capability of our method, we conducted experiments on the publicly available MMFi dataset [29]. MMFi was collected using TP-Link N750 WiFi APs and provides 3D joint annotations for 27 different actions performed by 40 subjects. We followed the official Setting 1 under Protocol 3, where all action samples are randomly divided into training and testing sets at a 3:1 ratio.

### B. Experimental Deployment

To assess the effectiveness of pose estimation, we adopt three commonly used evaluation metrics: Percentage of Correct Keypoints (PCK) [35], Mean Per Joint Position Error (MPJPE), and Procrustes Aligned MPJPE (PA-MPJPE) [37], all measured in millimeters. PCK measures the proportion of predicted keypoints falling within a specific threshold distance from the ground truth, typically denoted as PCK@$\alpha$, and is commonly used to assess local accuracy. MPJPE calculates the average Euclidean distance between predicted and true joint coordinates, reflecting the absolute positional accuracy. PA-MPJPE applies Procrustes alignment to the predictions and ground truth by removing translation, rotation, and scaling effects, allowing for a more accurate assessment of the structural similarity between the predicted and actual body configurations.

All experiments were conducted using the PyTorch framework on a single NVIDIA GeForce RTX 4090 GPU. Model validation was primarily performed on our self-collected 2D WiFi-based human pose dataset, with the number of training epochs set to 100. For the MMFi dataset, training was carried out for 50 epochs following the official settings.

For both datasets, the Adam optimizer was adopted with an initial learning rate of 0.0001. The batch size was set to 128 for training and 32 for testing. On the MMFi dataset, a StepLR learning rate scheduler was employed to enhance convergence, with a step size of 10 and a decay factor of 0.85.

### C. Results

We first evaluate the performance of VST-Pose on our self-collected dataset. In the experiments, the sliding window size was set to 3 frames with a stride of 2. The ViSTA-Former

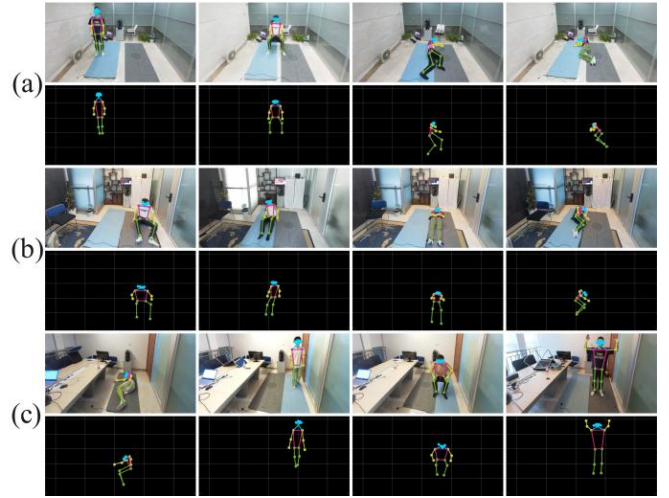

Fig. 7. Comparison of VST-Pose pose estimation (bottom) and ground truth (top). (a), (b), and (c) show the results from three different rooms.

### TABLE IV
#### VST-POSE PERFORMANCE WITH DIFFERENT WINDOW SIZES

| window | PCK@50 | PCK@20 | MPJPE | PA-MPJPE |
|---|---|---|---|---|
| 7 | 90.66 | 67.83 | 9.04 | 5.09 |
| 5 | 91.42 | 70.48 | 8.48 | 4.81 |
| 4 | 91.28 | 68.64 | 8.46 | 5.04 |
| 3 | 92.25 | 71.90 | 8.06 | 4.70 |

### TABLE V
#### COMPARISON OF PERFORMANCE BY STACKING DIFFERENT DST-BLOCK LAYERS IN THE VISTA-FORMER BACKBONE.

| Layer | PCK@50 | PCK@20 | MPJPE | PA-MPJPE |
|---|---|---|---|---|
| 1 | 90.08 | 63.29 | 9.60 | 5.46 |
| 3 | 91.87 | 70.17 | 8.44 | 4.87 |
| 5 | 92.25 | 71.90 | 8.06 | 4.70 |
| 7 | 92.07 | 72.05 | 8.18 | 4.73 |

backbone consists of five stacked DST-Blocks. The keypoint-wise detection performance is presented in Table I. The model achieves an overall mean PCK@50 of 92.2%, demonstrating high estimation accuracy. Notably, joints on the torso and lower limbs (e.g., knees and ankles) exhibit higher accuracy due to their larger motion amplitude and less articulated movement, while joints such as elbows and wrists show relatively lower precision. This discrepancy is attributed to the more complex motion patterns of distal joints, which pose greater challenges for learning.

Subsequently, we use the PCK@20 metric to analyze the performance across different action categories. As illustrated in Fig. 6, for static actions with stable postures such as "sitting" and "lying," the PCK@20 scores exceed 80%, indicating strong keypoint localization accuracy. In contrast, dynamic actions like "turning over" and "walking" exhibit slightly lower accuracy owing to rapid motion transitions and self-occlusion effects, primarily due to their more complex motion patterns and frequent limb occlusions, which pose greater challenges for precise keypoint detection.

To further illustrate the effectiveness of our system, we visualize the estimated poses alongside the ground truth annotations. As shown in Fig. 7, VST-Pose demonstrates accurate and temporally consistent 2D pose estimation.

| Velocity Method | PCK@50 | PCK@20 | MPJPE | PA-MPJPE |
|---|---|---|---|---|
| Without Velocity Modeling | 91.54 | 69.40 | 8.59 | 5.05 |
| With Velocity Modeling | 92.25 | 71.90 | 8.06 | 4.70 |

| Velocity Branch Source | PCK@50 | PCK@20 | MPJPE | PA-MPJPE |
|---|---|---|---|---|
| ST | 91.87 | 70.50 | 8.41 | 4.85 |
| TS+ST | 92.06 | 71.77 | 8.23 | 4.75 |
| TS | 92.25 | 71.90 | 8.06 | 4.70 |

| Fusion Strategy | PCK@50 | PCK@20 | MPJPE | PA-MPJPE |
|---|---|---|---|---|
| Without Velocity Fusion | 91.68 | 69.71 | 8.46 | 4.92 |
| With Velocity Fusion | 92.25 | 71.90 | 8.06 | 4.70 |

### D. Model Comparison

**Comparison on self-collected dataset.** We conducted comparative experiments on our self-collected dataset against existing WiFi HPE methods, as shown in Table II. The results demonstrate that VST-Pose outperforms existing approaches across all evaluation metrics. The notable improvement in PCK@20 highlights the model's enhanced capability in fine-grained keypoint localization.

**Comparison on MMFI dataset.** Given that our dataset only includes 2D keypoints, we further test VST-Pose on publicly available MMFI dataset, which contains 3D human keypoints. We set the sliding window size to 10 frames with a stride of 3. Considering the higher WiFi resolution in MMFI, the ViSTA-Former backbone was reduced to one layer to improve model stability and training efficiency. All models were trained for 50 epochs under consistent settings. As shown in Table III, significant improvements in PCK@20 and MPJPE demonstrate the model's ability to track more precise keypoint locations, and provide better 3D human skeleton construction.

### E. Ablation Study

We conducted a series of ablation experiments on our self-collected dataset, analyzing the impact of key design choices on model performance.

**Sliding Window Size Selection.** Our method adopts a multi-frame-to-multi-frame temporal modeling strategy, where CSI sequences are segmented using a sliding window. We compared temporal windows of 3, 4, 5, and 7 frames, with a fixed stride of 2 to reduce temporal redundancy. As shown in Table IV The results indicate that a window size of 3 performs the best. Although shorter, this window allows the model to better capture subtle motion variations, leading to superior precision in fine-grained coordinate estimation compared to single-frame models.

**Network Depth Selection.** The ViSTA-Former backbone is Table IV composed of stacked DST-Blocks. We experimented with 1, 3, 5, and 7 layers of DST-Blocks. Table V shows consistent improvements in PCK and PA-MPJPE as the depth increases up to 5 layers. Beyond this point, performance gains plateau, suggesting diminishing returns. Considering the trade-off between computational cost and accuracy, we adopt a 5-layer configuration as the default for our self-collected dataset.

**Impact of Velocity Modeling.** As shown in Table VI, incorporating velocity modeling improves PCK@20 by 2.5%, indicating its effectiveness in enhancing fine-grained keypoint tracking. Table VII further examines the source of velocity features, where using the TS branch of the DST-Block yields superior results, highlighting its specialized temporal modeling capability within the dual-stream design. Table VIII analyzes the late fusion strategy, showing that the dynamic features extracted by the velocity branch effectively compensate for the backbone's limited motion awareness across frames.

### F. Limitations

Although VST-Pose demonstrates superior performance in WiFi-based pose estimation, several limitations remain. First, the current system requires synchronized acquisition of CSI and video data, resulting in a complex and costly deployment process. Second, the current setup uses a single transmitter–receiver pair in one direction, which limits spatial resolution. Lastly, the current method struggles with cross-domain modeling, highlighting the need for more diverse data collection or the adoption of domain adaptation techniques. These challenges present promising directions for future work.

## VI. CONCLUSION

This paper presents VST-Pose, a novel attention-based network for WiFi-based human pose estimation that integrates explicit spatiotemporal modeling with velocity dynamics . By designing the ViSTA-Former backbone with dual-stream spatiotemporal attention and a dedicated velocity branch, our method effectively captures both structural and dynamic cues from CSI signals. We evaluate the proposed model on both a self-collected 2D dataset and the public 3D dataset MMFI. Experimental results demonstrate that VST-Pose consistently outperforms existing approaches across multiple evaluation metrics. These achievements lay the foundation for practical applications in home care and privacy-sensitive environments such as non-intrusive health monitoring and fall detection.

## REFERENCES

[1] A. Toshev and C. Szegedy, "DeepPose: Human pose estimation via deep neural networks," in *Proc. IEEE Conf. Comput. Vis. Pattern Recognit. (CVPR)*, Columbus, OH, USA, Jun. 2014, pp. 1653–1660.

[2] H.-S. Fang *et al.*, "AlphaPose: Whole-body regional multi-person pose estimation and tracking in real-time," 2022, *arXiv*: 2211.03375.

[3] Y. Chen, Z. Wang, Y. Peng, Z. Zhang, G. Yu, and J. Sun, "Cascaded pyramid network for multi-person pose estimation," in *Proc. IEEE Conf. Comput. Vis. Pattern Recognit. (CVPR)*, Salt Lake City, Jun. 2018, pp. 7103–7112.

[4] Z. Cao, G. Hidalgo, T. Simon, S. E. Wei, and Y. Sheikh, "OpenPose: Realtime multi-person 2D pose estimation using part affinity fields," IEEE Trans. Pattern Anal. Mach. Intell., vol. 43, no. 1, pp. 172–186, Jan. 2021.

[5] S. Lee *et al.*, "Human pose estimation in extremely low-light conditions," in *Proc. IEEE Conf. Comput. Vis. Pattern Recognit. (CVPR)*, Jun. 2023, pp. 704–714.

[6] W. Yang, W. Ouyang, X. Wang, J. Ren, H. Li, and X. Wang, "3D Human pose estimation in the wild by adversarial learning," in *Proc. IEEE Conf. Comput. Vis. Pattern Recognit. (CVPR)*, Jun. 2018, pp. 5255–5264.

[7] Y. Ma, G. Zhou, and S. Wang, "WiFi sensing with channel state information: A survey," *ACM Comput. Surv.*, vol. 52, no. 3, pp. 1–36, Jul. 2020.

[8] K. Qian, C. Wu, Z. Zhou, Y. Zheng, Z. Yang, and Y. Liu, "Inferring motion direction using commodity Wi-Fi for interactive exergames," in *Proc. CHI Conf. Human Factors Comput. Syst. (CHI)*, Denver, CO, USA, May 2017, pp. 1961–1972.

[9] Y. Wang, J. Liu, Y. Chen, M. Gruteser, J. Yang, and H. Liu, "E-eyes: Device-free location-oriented activity identification using fine-grained Wi-Fi signatures," in *Proc. 20th Annu. Int. Conf. Mobile Comput. Netw. (MobiCom)*, Maui, HI, USA, Sep. 2014, pp. 617–628.

[10] Y. Zhang *et al.*, "Widar3.0: Zero-effort cross-domain gesture recognition with Wi-Fi," *IEEE Trans. Pattern Anal. Mach. Intell.*, vol. 44, no. 11, pp. 8671–8688, Nov. 2022.

[11] N. Bao *et al.*, "Wi-Breath: A WiFi-based contactless and real-time respiration monitoring scheme for remote healthcare," *IEEE J. Biomed. Health Inform.*, vol. 27, no. 5, pp. 2276–2285, May 2023.

[12] S. Yousefi, H. Narui, S. Dayal, S. Ermon, and S. Valaee, "A Survey on behavior recognition Using WiFi Channel State Information," *IEEE Commun. Mag.*, vol. 55, no. 10, pp. 98–104, Oct. 2017.

[13] J. Yang, X. Chen, H. Zou, D. Wang, Q. Xu, and L. Xie, "EfficientFi: Toward Large-scale lightweight WiFi Sensing via CSI Compression," *IEEE Internet Things J.*, vol. 9, no. 15, pp. 13086–13095, Aug. 2022.

[14] F. Wang, S. Panev, Z. Dai, J. Han, and D. Huang, "Can WiFi estimate person pose?" 2019, arXiv:1904.00277.

[15] L. Guo, Z. Lu, X. Wen, S. Zhou, and Z. Han, "From signal to image: Capturing fine-grained human poses With Commodity Wi-Fi," *IEEE Commun. Lett.*, vol. 24, no. 4, pp. 802–806, Apr. 2020.

[16] Y. Zhou, A. Zhu, C. Xu, F. Hu, and Y. Li, "PerUnet: Deep signal channel attention in unet for WiFi-based human pose estimation," *IEEE Sens. J.*, vol. 22, no. 20, pp. 19750–19760, Oct. 2022.

[17] W. Jiang *et al.*, "Towards 3D human pose construction using WiFi," in *Proc. 26th Annu. Int. Conf. Mobile Comput. Netw.*, 2020, pp. 1–14.

[18] Y. Wang, L. Guo, Z. Lu, X. Wen, S. Zhou, and W. Meng, "From point to space: 3D moving human pose estimation using commodity WiFi," *IEEE Commun. Lett.*, vol. 25, no. 7, pp. 2235–2239, Jul. 2021.

[19] Y. Ren, Z. Wang, Y. Wang, S. Tan, Y. Chen, and J. Yang, "GoPose: 3D human pose estimation using WiFi," *Proc. ACM Interact. Mob. Wearable Ubiquitous Technol.*, vol. 6, no. 2, pp. 1–25, Jul. 2022.

[20] F. Wang, S. Zhou, S. Panev, J. Han, and D. Huang, "Person-in-WiFi: Fine-grained person perception using WiFi," in *Proc. IEEE/CVF Int. Conf. Comput. Vis.*, Oct. 2019, pp. 5452–5461.

[21] K. Yan, F. Wang, B. Qian, H. Ding, J. Han, and X. Wei, "Person-in-WiFi 3D: End-to-end multi-person 3D pose estimation with Wi-Fi," in *Proc. IEEE Conf. Comput. Vis. Pattern Recognit. (CVPR)*, Seattle, WA, USA, Jun. 2024, pp. 969–978.

[22] Geng J., Huang D., and Torre F. D., "DensePose from WiFi," 2022, *arXiv*: 2301.00250.

[23] Y. Chen, J. Guo, S. Guo, J. Zhou, and D. Tao, "Towards robust and Rrealistic human pose estimation via WiFi signals," 2025, *arXiv*: 2501.09411.

[24] Y. Zhou, J. Yang, H. Huang, and L. Xie, "AdaPose: Toward cross-site device-free human pose estimation with commodity Wi-Fi," *IEEE Internet Things J.*, vol. 11, no. 24, pp. 40255–40267, Dec. 2024.

[25] Y. Ren, Z. Wang, S. Tan, Y. Chen, and J. Yang, "Winect: 3D human pose tracking for free-form activity using commodity WiFi," *Proc. ACM Interact. Mob. Wearable Ubiquitous Technol.*, vol. 5, no. 4, pp. 1–29, Dec. 2021.

[26] T.-W. Hsu and H.-Y. Hsieh, "Robust multi-user pose estimation based on spatial and temporal features from Wi-Fi CSI," in *Proc. IEEE Int. Conf. Commun. (ICC)*, Denver, CO, USA, Jun. 2024, pp. 1600–1605.

[27] A. Vaswani et al., "Attention is all you need," 2017, arXiv:1706.03762.

[28] W. Zhu, X. Ma, Z. Liu, L. Liu, W. Wu, and Y. Wang, "MotionBERT: A unified perspective on learning human motion representations," in *Proc. IEEE Int. Conf. Comput. Vis. (ICCV)*, Oct. 2023, pp. 15039–15053.

[29] J. Yang *et al.*, "MM-Fi: Multi-modal non-intrusive 4D human dataset for versatile wireless sensing," 2023, *arXiv*: 2305.10345.

[30] S. Z. Homayounfar and T. L. Andrew, "Wearable sensors for monitoring human motion: A review on mechanisms, materials, and challenges," *SLAS Technol.*, vol. 25, no. 1, pp. 9–24, Feb. 2020.

[31] J. Li *et al.*, "LiDARCap: Long-range markerless 3D human motion capture with LiDAR point clouds," in *Proc. IEEE Conf. Comput. Vis. Pattern Recognit. (CVPR)*, New Orleans, LA, USA, Jun. 2022, pp. 20470–20480.

[32] M. Zhao et al., "Through-wall human pose estimation using radio signals," in *Proc. IEEE Conf. Comput. Vis. Pattern Recognit.*, 2018, pp. 7356–7365.

[33] M. Zhao et al. , "RF-based 3-D skeleton,"in *Proc. ACM SIGCOMM Conf. Data Commun. (SIGCOMM)*, Aug. 2018, pp. 267–281

[34] Y. Zhou, C. Xu, L. Zhao, A. Zhu, F. Hu, and Y. Li, "CSI-Former: Pay more attention to pose estimation with WiFi," *Entropy*, vol. 25, no. 1, p. 20, Dec. 2022.

[35] Y. Zhou, H. Huang, S. Yuan, H. Zou, L. Xie, and J. Yang, "MetaFi++: WiFi-enabled transformer-based human pose estimation for metaverse avatar simulation," *IEEE Internet Things J.*, vol. 10, no. 16, pp. 14128–14136, Aug. 2023.

[36] D. Halperin, W. Hu, A. Sheth, and D. Wetherall, "Tool Release: Gathering 802.11n Traces With Channel State Information," *ACM SIGCOMM Comput. Commun. Rev.*, vol. 41, no. 1, p. 53, Jan. 2011.

[37] T. D. Gian, T. D. Lai, T. V. Luong, K.-S. Wong, and V.-D. Nguyen, "HPE-Li: Wi-Fi-enabled lightweight dual selective kernel convolution for human pose estimation," in *Proc. Eur. Conf. Comput. Vis. (ECCV)*, vol. 15089. Cham, Switzerland, Jan. 2025, pp. 93–111.